\documentclass[11pt,a4paper]{article}
\usepackage[hyperref]{acl2021}
\usepackage{times}
\usepackage{latexsym}

\usepackage{microtype}

\usepackage{amsmath,amsfonts,bm}
\usepackage{microtype}
\usepackage{graphicx}
\usepackage{booktabs} 
\usepackage{multirow}
\usepackage{subcaption}
\usepackage{marginnote}
\usepackage{bbm}
\usepackage{float}
\usepackage[shortlabels]{enumitem}
\usepackage{tikz}
\usepackage{ifthen}
\usepackage{stmaryrd}
\usepackage{wrapfig}
\usepackage{caption}
\usepackage{soul}
\usepackage{pifont}
\usepackage{nicefrac}
\usepackage{adjustbox}
\usepackage{cancel}
\usepackage{inconsolata}
\usepackage[normalem]{ulem}

\aclfinalcopy 
\def\aclpaperid{3325} 

\newif\ifcomments
\commentstrue
\ifcomments
    \providecommand\matt[1]{\textcolor{teal}{[Matt: #1]}}
    \providecommand\nitish[1]{\textcolor{violet}{[Nitish: {#1}]}}
    \providecommand\sameer[1]{\textcolor{purple}{[Sameer: #1]}}
\else
    \providecommand{\matt}[1]{}
    \providecommand{\nitish}[1]{}
    \providecommand{\sameer}[1]{}
\fi

\newcommand{\utterance}[1]{\emph{#1}}
\newcommand{\program}[1]{\textcolor{teal!75!black}{\texttt{#1}}}
\newcommand{\denotation}[1]{\llbracket #1 \rrbracket}
\newcommand{\action}[2]{$\program{#1} \rightarrow \program{#2}$}

\title{Enforcing Consistency in Weakly Supervised Semantic Parsing}

\author{
  Nitish Gupta\thanks{\hspace{0.3em} Work done while interning with Allen Institute for AI.} \\
  University of Pennsylvania \\
  \texttt{nitishg@seas.upenn.edu} \\\And
  Sameer Singh \\
  University of California, Irvine \\
  \texttt{sameer@uci.edu} \\\And
  Matt Gardner \\
  Allen Institute for AI \\
  \texttt{mattg@allenai.org} \\
  }

\date{}

\begin{document}
\maketitle
\begin{abstract}
The predominant challenge in weakly supervised semantic parsing is that of \emph{spurious programs} that evaluate to correct answers for the wrong reasons. Prior work uses elaborate search strategies to mitigate the prevalence of spurious programs; however, they typically consider only one input at a time. In this work we explore the use of consistency between the output programs for related inputs to reduce the impact of spurious programs. We bias the program search (and thus the model's training signal) towards programs that map the same phrase in related inputs to the same sub-parts in their respective programs. 
Additionally, we study the importance of designing logical formalisms that facilitate this kind of consistency-based training. We find that a more consistent formalism leads to improved model performance even without consistency-based training. When combined together, these two insights lead to a 10\% absolute improvement over the best prior result on the Natural Language Visual Reasoning dataset.
\end{abstract}

\section{Introduction}
Semantic parsers map a natural language utterance into an executable meaning representation, called a logical form or program~\citep{zelle1996, zettlemoyer2005}. These programs can be executed against a context (e.g., database, image, etc.) to produce a denotation (e.g., answer) for the input utterance.
Methods for training semantic parsers from only (utterance, denotation) supervision have been developed~\citep{clarke2010, Liang2011, berant2013}; however, training from such weak supervision is challenging. The parser needs to search for the correct program from an exponentially large space, and the presence of \emph{spurious programs}---incorrect representations that evaluate to the correct denotation---greatly hampers learning.
Several strategies have been proposed to mitigate this issue~\cite{GuuFromLT2017, LiangMAPO2018, DasigiIterativeSF2019}.
Typically these approaches consider a single input utterance at a time and explore ways to score programs.

\begin{figure}
\centering
\includegraphics[width=\columnwidth]{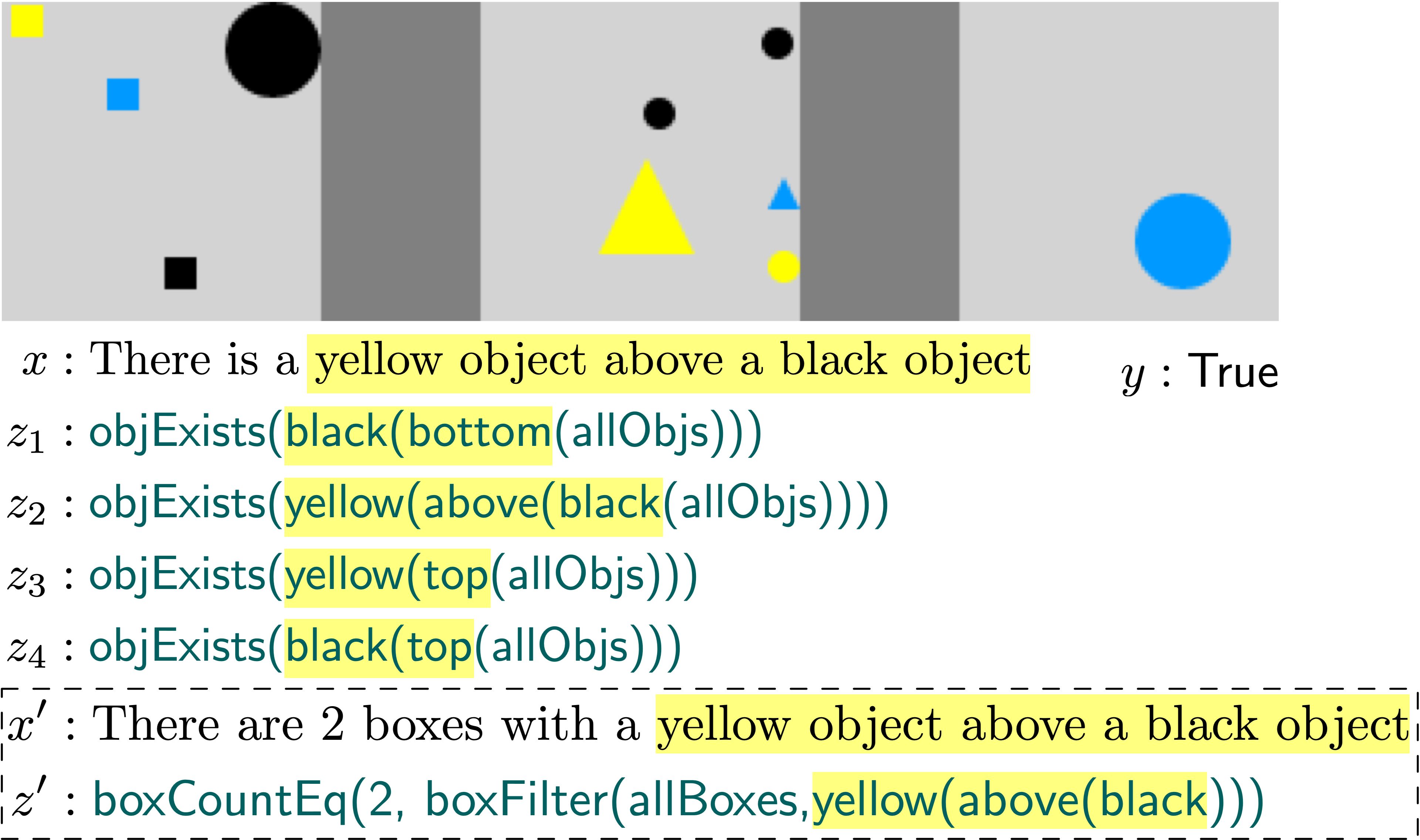}
\caption{\label{fig:overview}
Utterance $x$ and its program candidates $z_1$-$z_4$, all of which evaluate to the correct denotation (\texttt{True}).
$z_2$ is the correct interpretation; other programs are \emph{spurious}.
Related utterance $x'$ shares the phrase \utterance{yellow object above a black object} with $x$.  
Our consistency reward would score $z_2$ the highest since it maps the shared phrase most similarly compared to $z'$.
}
\end{figure}

In this work we encourage consistency between the output programs of related natural language utterances to mitigate the issue of spurious programs. 
Consider related utterances, \utterance{There are two boxes with three yellow squares} and \utterance{There are three yellow squares}, both containing the phrase \utterance{three yellow squares}.
Ideally, the correct programs for the utterances should contain similar sub-parts that corresponds to the shared phrase.
To incorporate this intuition during search, we propose a consistency-based reward to encourage programs for related utterances that share sub-parts corresponding to the shared phrases (\S\ref{sec:consistency-reward}). 
By doing so, the model is provided with an additional training signal to distinguish between programs based on their consistency with programs predicted for related utterances. 

We also show the importance of designing the logical language in a manner such that the ground-truth programs for related utterances are consistent with each other. 
Such consistency in the logical language would facilitate the consistency-based training proposed above, and encourage the semantic parser to learn generalizable correspondence between natural language and program tokens. 
In the previously proposed language for the Natural Language Visual Reasoning dataset \cite[NLVR;][]{SuhrNLVR2017}, we notice that the use of macros leads to inconsistent interpretations of a phrase depending on its context. We propose changes to this language such that a phrase in different contexts can be interpreted by the same program parts (\S\ref{sec:language}). 

We evaluate our proposed approaches on NLVR using the semantic parser of \citet{DasigiIterativeSF2019} as our base parser. 
On just replacing the old logical language for our proposed language we see an 8\% absolute improvement in consistency, the evaluation metric used for NLVR (\S\ref{sec:exp}).
Combining with our consistency-based training leads to further improvements; overall 10\% over the best prior model, 
reporting a new state-of-the-art on the NLVR dataset.

\section{Background}
\label{sec:background}
In this section we provide a background on the NLVR dataset~\cite{SuhrNLVR2017} and the semantic parser of \citet{DasigiIterativeSF2019}.

\paragraph{Natural Language Visual Reasoning (NLVR)} dataset contains human-written natural language utterances, where each utterance is paired with 4 synthetically-generated images. Each (utterance, image) pair is annotated with a binary truth-value denotation denoting whether the utterance is true for the image or not. 
Each image is divided into three \emph{boxes}, where each box contains 1-8 \emph{objects}. Each object has four properties: \emph{position} (x/y coordinates), \emph{color} (black, blue, yellow), \emph{shape} (triangle, square, circle), and \emph{size} (small, medium, large). 
The dataset also provides a structured representation of each image which we use in this paper. 
Figure~\ref{fig:overview} shows an example from the dataset.

\paragraph{Weakly supervised iterative search parser}
We use the semantic parser of \citet{DasigiIterativeSF2019} which is a grammar-constrained encoder-decoder with attention model from ~\citet{WikiTablesParser2017}.
It learns to map a natural language utterance $x$ into a program $z$ such that it evaluates to the correct denotation $y = \denotation{z}^{r}$ when executed against the structured image representation~$r$. 
\citet{DasigiIterativeSF2019} use a manually-designed, typed, variable-free, functional query language for NLVR, inspired by the GeoQuery language~\cite{zelle1996}. 

Given a dataset of triples ($x_i$, $c_i$, $y_i$), where $x_i$ is an utterance, $c_i$ is the set of images associated to it, and $y_i$ is the set of corresponding denotations, their approach iteratively alternates between two phases to train the parser: Maximum marginal likelihood (MML) and a Reward-based method (RBM). 
In MML, for an utterance $x_i$, the model maximizes the marginal likelihood of programs in a given set of logical forms $Z_i$, all of which evaluate to the correct denotation.
The set $Z_i$ is constructed either by performing a heuristic search, or generated from a trained semantic parser.

The reward-based method maximizes the (approximate) expected value of a reward function $\mathcal{R}$.
\begin{equation}
\label{eq:rbm}
    \max_{\theta} \sum_{\forall i} \mathbb{E}_{\tilde{p}(z_i|x_i; \theta)} \mathcal{R}(x_i, z_i, c_i, y_i)
\end{equation}
Here, $\tilde{p}$ is the \emph{re-normalization} of the probabilities assigned to the programs on the beam, and the reward function $\mathcal{R} = 1$ if $z_i$ evaluates to the correct denotation for all images in $c_i$, or $0$ otherwise. Please refer \citet{DasigiIterativeSF2019} for details.

\section{Consistency reward for programs}
\label{sec:consistency-reward}
Consider the utterance $x$ = \utterance{There is a yellow object above a black object} in Figure~\ref{fig:overview}. There are many program candidates decoded in search that evaluate to the correct denotation. Most of them are \emph{spurious}, i.e., they do not represent the meaning of the utterance and only coincidentally evaluate to the correct output. The semantic parser is expected to distinguish between the correct program and spurious ones by identifying correspondence between parts of the utterance and the program candidates. 
Consider a related utterance $x'$ = \utterance{There are 2 boxes with a yellow object above a black object}. The parser should prefer programs for $x$ and $x'$ which contain similar sub-parts corresponding to the shared phrase $p$~=~\utterance{yellow object above a black object}. That is, the parser should be consistent in its interpretation of a phrase in different contexts. 
To incorporate this intuition during program search, we propose an additional reward to programs for an utterance that are consistent with programs for a related utterance. 

Specifically, consider two related utterances $x$ and $x'$ that share a phrase $p$.
We compute a reward for a program candidate $z$ of $x$ based on how similarly it maps the phrase $p$ as compared to a program candidate $z'$ of $x'$.  To compute this reward we need (a)~\emph{relevant program parts} in $z$ and $z'$ that correspond to the phrase $p$, and (b)~a \emph{consistency reward} that measures consistency between those parts.

\paragraph{(a) Relevant program parts}
Let us first see how to identify relevant parts of a program $z$ that correspond to a phrase $p$ in the utterance.

Our semantic parser (from ~\citet{WikiTablesParser2017}) outputs a linearized version of the program $z = [z^1, \ldots, z^T]$, decoding one action at a time from the logical language. At each time step, the parser predicts a normalized attention vector over the tokens of the utterance, denoted by $[a^t_1, \ldots, a^t_N]$ for the $z^t$ action. Here, $\sum_{i=1}^{N}a^t_i = 1$ and $a^t_i \geq 0$ for $i \in [1, N]$.
We use these attention values as a relevance score between a program action and the utterance tokens. Given the phrase $p$ with token span $[m, n]$, we identify the relevant actions in $z$ as the ones whose total attention score over the tokens in $p$ exceeds a heuristically-chosen threshold~$\tau = 0.6$.
\begin{equation}
    A(z, p) = \Big\{z^t \; \big| \; t \in [1, T] \; \text{and} \; \sum_{i=m}^{n} a^t_i \geq \tau\Big\}
\end{equation}
This set of program actions $A(z, p)$ is considered to be generated due to the phrase $p$. For example, for utterance \utterance{There is a yellow object above a black object}, with program \program{objExists(yellow(above(black(allObjs)))}, this approach could identify that for the phrase \utterance{yellow object above a black object} the actions corresponding to the functions \program{yellow}, \program{above}, and \program{black} are relevant.

\paragraph{(b) Consistency reward}
Now, we will define a reward for the program $z$ based on how consistent its mapping of the phrase $p$ is w.r.t. the program $z'$ of a related utterance.
Given a related program $z'$ and its relevant action set $A(z', p)$, we define the consistency reward $S(z, z', p)$ as the F1 score for the action set $A(z, p)$ when compared to $A(z', p)$. 
If there are multiple shared phrases $p_i$ between $x$ and $x'$, we can compute a weighted average of different $S(z, z', p_i)$ to compute a singular consistency reward $S(z, z')$ between the programs $z$ and $z'$. In this work, we only consider a single shared phrase $p$ between the related utterances, hence $S(z, z', p) = S(z, z', p)$ in our paper.

As we do not know the gold program for $x'$, we decode top-K program candidates using beam-search and discard the ones that do not evaluate to the correct denotation. We denote this set of programs by $Z'_c$.
Now, to compute a consistency reward $\mathcal{C}(x, z, x')$ for the program $z$ of $x$,we take a weighted average of $S(z, z')$ for different $z' \in Z'_c$ where the weights correspond to the probability of the program $z'$ as predicted by the parser.
\begin{equation}
\mathcal{C}(x, z, x') = \sum_{z' \in Z'_c} \tilde{p}(z'|x'; \theta) S(z, z') 
\end{equation}

\paragraph{Consistency reward based parser} Given $x$ and a related utterance $x'$, we use $\mathcal{C}(x, z, x')$ as an additional reward in Eq.~\ref{eq:rbm} to upweight programs for $x$ that are consistent with programs for $x'$. 
\begin{equation*}
\label{eq:paired-rbm}
\max_{\theta} \sum_{\forall i} \mathbb{E}_{\tilde{p}(z_i|x_i; \theta)} \big[ \mathcal{R}(x_i, z_i, c_i, y_i) + \mathcal{C}(x_i, z_i, x'_i) \big]
\end{equation*}
This consistency-based reward pushes the parser's probability mass towards programs that have consistent interpretations across related utterances, thus providing an additional training signal over simple denotation accuracy.
The formulation presented in this paper assumes that there is a single related utterance $x'$ for the utterance $x$. If multiple related utterances are considered, the consistency reward $\mathcal{C}(x, z, x'_j)$ for different related utterances $x'_j$ can be summed/averaged to compute a single consistency reward $\mathcal{C}(x, z)$  the program $z$ of utterance $x$ based on all the related utterances.
 
\begin{table*}
\centering
\begin{tabular}{l cc cc cc}
\toprule
\bf \multirow{2}[3]{*}{Model} & \multicolumn{2}{c}{\bf Dev} &  \multicolumn{2}{c}{\bf Test-P} & \multicolumn{2}{c}{\bf Test-H} \\
\cmidrule(lr){2-3}  \cmidrule(lr){4-5} \cmidrule(lr){6-7}
& \bf Acc.  & \bf Cons. & \bf Acc.  & \bf Cons. & \bf Acc.  & \bf Cons.  \\
\midrule
\textsc{Abs. Sup.}~\cite{GoldmanWeaklySSP2018}              & 84.3  & 66.3 & 81.7  & 60.1 & -    & -   \\
\textsc{Abs. Sup. + ReRank}~\cite{GoldmanWeaklySSP2018}     & 85.7  & 67.4 & 84.0  & 65.0 & 82.5 & 63.9   \\
\textsc{Iterative Search}~\cite{DasigiIterativeSF2019}      & 85.4  & 64.8 & 82.4  & 61.3 & 82.9 & 64.3   \\
\addlinespace[1mm]
\; + Logical Language Design (ours)   & 88.2  & 73.6 & 86.0  & 69.6 & -    & -       \\
\; + Consistency Reward (ours)         & \textbf{89.6}  & \textbf{75.9} & \textbf{86.3}  & \textbf{71.0} & \textbf{89.5} & \textbf{74.0}   \\
\bottomrule
\end{tabular}
\centering
    \caption{ \label{tab:mainresults}
    \textbf{Performance on NLVR:} Design changes in the logical language and consistency-based training, both significantly improve performance. Larger improvements in consistency indicate that our approach efficiently tackles spurious programs.}
\end{table*}

\section{Consistency in Language}
\label{sec:language}
The consistency reward (\S\ref{sec:consistency-reward}) makes a key assumption about the logical language in which the utterances are parsed: that the gold programs for utterances sharing a natural language phrase actually correspond to each other.
For example, that the phrase \utterance{yellow object above a black object} would always get mapped to \program{yellow(above(black))} irrespective of the utterance it occurs in.

On analyzing the logical language of \citet{DasigiIterativeSF2019}, we find that this assumption does not hold true. Let us look at the following examples:
\\
$x_1$: \utterance{There are items of at least two different colors}\\
$z_1$: \program{objColorCountGrtEq(2, allObjs)}\\
$x_2$: \utterance{There is a box with items of at least two different colors}\\
$z_2$: \program{boxExists(} \\
\hphantom{~~~~~~}\program{memberColorCountGrtEq(2,} \program{allBoxes))}\\
Here the phrase \utterance{items of at least two different colors} is interpreted differently in the two utterances. In $x_2$, a macro function \program{memberColorCountGrtEq} is used, which internally calls \program{objColorCountGrtEq} for each \emph{box} in the image.
Now consider,\\
$x_3$: \utterance{There is a tower with exactly one block}\\
$z_3$: \program{boxExists(memberObjCountEq(1,allBoxes))} \\
$x_4$: \utterance{There is a tower with a black item on the top}\\
$z_4$: \program{objExists(black(top(allObjs)))}\\
Here the phrase \utterance{There is a tower} is interpreted differently: $z_3$ uses a macro for filtering boxes based on their object count and interprets the phrase using \program{boxExists}. In the absence of a complex macro for checking \utterance{black item on the top}, $z_4$ resorts to using \program{objExists} making the interpretation of the phrase inconsistent.
These examples highlight that these macros, while they shorten the search for programs, make the language inconsistent.

We make the following changes in the logical language to make it more consistent. Recall from \S\ref{sec:background} that each NLVR image contains 3 boxes each of which contains 1-8 objects.
We remove macro functions like \program{memberColorCountGrtEq}, and introduce a generic \program{boxFilter} function. This function takes two arguments, a set of \emph{boxes} and a filtering function \program{f}: $\text{\program{Set[Obj]}} \rightarrow \text{\program{bool}}$, and prunes the input set of boxes to the ones whose objects satisfies the filter \texttt{f}. By doing so, our language is able to reuse the same object filtering functions across different utterances. In this new language, the gold program for the utterance $x_2$
would be \\
$z_2$: \program{boxCountEq(1, boxFilter(allBoxes,} \\
\hphantom{~~~~~~~~~~~~}\program{objColorCountGrtEq(2)))} \\    
By doing so, our logical language can now consistently interpret the phrase \utterance{items of at least two different colors} using the object filtering function \texttt{f}: \program{objColorCountGrtEq(2)} across both $x_1$ and $x_2$.
Similarly, the gold program for $x_4$
in the new logical language would be \\
$z_4$: \program{boxExists(boxFilter(allBoxes, black(top)))} \\
making the interpretation of \utterance{There is a box} consistent with $x_3$.
Please refer appendix \S\ref{app:language} for details.

\section{Experiments}
\label{sec:exp}

\paragraph{Dataset} 
We report results on the standard development, public-test, and hidden-test splits of NLVR. 
The training data contains 12.4k (utterance, image) pairs where each of 3163 utterances are paired with $4$ images. 
Each evaluation set roughly contains 270 unique utterances.

\paragraph{Evaluation Metrics} (1)~\emph{Accuracy} measures the proportion of examples for which the correct denotation is predicted. (2)~
Since each utterance in NLVR is paired with 4 images, a \emph{consistency} metric is used, which measures the proportion of utterances for which the correct denotation is predicted for all associated images. 
Improvement in this metric is indicative of correct program prediction as it is unlikely for a spurious program to correctly make predictions on multiple images.

\paragraph{Experimental details} We use the same parser, training methodology, and hyper-parameters as \citet{DasigiIterativeSF2019}. For discovering related utterances, we manually identify $\sim$10 sets of equivalent phrases that are common in NLVR. For example, \utterance{there are \emph{NUM} boxes}, \utterance{\emph{COLOR1} block on a \emph{COLOR2} block}, etc. 
For each utterance that contains a particular phrase, we pair it with one other randomly chosen utterance that shares the phrase.
We make 1579 utterance pairs in total.
Refer appendix \S\ref{app:dataset} for details about data creation.\footnote{We release the data and code at  \url{https://www.github.com/nitishgupta/allennlp-semparse/tree/nlvr-v2/scripts/nlvr_v2}}

\paragraph{Baselines} We compare against the state-of-the-art models; \textsc{Abs. Sup.}~\cite{GoldmanWeaklySSP2018} that uses abstract examples, \textsc{Abs. Sup. + ReRank} that uses additional data and reranking, and the iterative search parser of ~\citet{DasigiIterativeSF2019}.

\paragraph{Results} Table~\ref{tab:mainresults} compares the performance of our two proposed methods to enforce consistency in the decoded programs with the previous approaches. We see that changing the logical language to a more consistent one (\S\ref{sec:language}) significantly improves performance: the accuracy improves by 2-4\% and consistency by 4-8\% on the dev. and public-test sets.
Additionally, training the parser using our proposed consistency reward (\S\ref{sec:consistency-reward}) further improves performance: accuracy improves by 0.3-0.4\% but the consistency significantly improves by 1.4-2.3\%.\footnote{We report average performance across 10 runs trained with different random seeds. All improvements in consistency are statistically significant (p-value $<$ 0.05) based on the stochastic ordering test~\cite{deepStatDror2019}.}
On the hidden-test set of NLVR, our final model improves accuracy by 7\% and consistency by 10\% compared to previous approaches.
Larger improvements in consistency across evaluation sets indicates that our approach to enforce consistency between programs of related utterances greatly reduces the impact of spurious programs.

\section{Conclusion}
We proposed two approaches to mitigate the issue of spurious programs in weakly supervised semantic parsing by enforcing consistency between output programs. First, a consistency based reward that biases the program search towards programs that map the same phrase in related utterances to similar sub-parts. Such a reward provides an additional training signal to the model by leveraging related utterances. Second, we demonstrate the importance of logical language design such that it facilitates such consistency-based training. The two approaches combined together lead to significant improvements in the resulting semantic parser.

\section*{Acknowledgement}
We would like to thank Pradeep Dasigi for helping us with the code for preprocessing NLVR and the Iterative Search model, Alane Suhr for getting us our model's evaluation results on the hidden test set in a timely manner, and the anonymous reviewers for their helpful comments. This work is supported in part by NSF award \#IIS-1817183.

\bibliographystyle{acl_natbib}
\bibliography{references}

\clearpage
\appendix

\section{Logical language details}
\label{app:language}
In Figure~\ref{fig:example-prog}, we show an example utterance with its gold program according to our proposed logical language. 
We use function composition and function currying to maintain the variable-free nature of our language. 
For example, action $z^7$ uses function composition to create a function from \program{Set[Object]} $\rightarrow$ \program{bool} by composing two functions, from \program{Set[Object]} $\rightarrow$ \program{bool} and \program{Set[Object]} $\rightarrow$ \program{Set[Object]}.
Similarly, action $z^{11}$ creates a function from \program{Set[Object]} $\rightarrow$ \program{Set[Object]} by composing two functions with the same signature.

Actions $z^8$ - $z^{10}$ use function currying to curry the 2-argument function \program{objectCountGtEq} by giving it one \texttt{int}=2 argument. This results in a 1-argument function \program{objectCountGtEq(2)} from \program{Set[Object]} $\rightarrow$ \program{bool}.

\begin{figure*}[h!]
\fbox{
\resizebox{\textwidth}{!}{
\begin{minipage}{\textwidth}
$x$: \utterance{There is one box with at least 2 yellow squares}\\
$z$: \program{boxCountEq(1, boxFilter(allBoxes, objectCountGtEq(2)(yellow(square))))} \\

Program actions for $z$: \\
$z^1$: \action{bool}{[$<$int,[Set[Box]:bool$>$, int, Set[Box]]} \\
$z^2$: \action{$<$int,[Set[Box]:bool$>$}{boxCountEq} \\
$z^3$: \action{int}{1} \\
$z^4$: \action{Set[Box]}{[$<$Set[Box],$<$Set[Object]:bool$>$:Set[Box]$>$, Set[Box], $<$Set[Object]:bool$>$]} \\
$z^5$: \action{$<$Set[Box],$<$Set[Object]:bool$>$:Set[Box]$>$}{boxFilter} \\
$z^6$: \action{Set[Box]}{allBoxes} \\
$z^7$: \action{$<$Set[Object]:bool$>$}{[*, $<$Set[Object]:bool$>$, $<$Set[Object]:Set[Object]$>$]} \\
$z^8$: \action{$<$Set[Object]:bool$>$}{[$<$int,Set[Object]:bool$>$, int]} \\
$z^9$: \action{$<$int,Set[Object]:bool$>$}{objectCountGtEq} \\
$z^{10}$: \action{int}{2} \\
$z^{11}$: \action{$<$Set[Object]:Set[Object]$>$}{[*, $<$Set[Object]:Set[Object]$>$, $<$Set[Object]:Set[Object]$>$]} \\
$z^{12}$: \action{$<$Set[Object]:Set[Object]$>$}{yellow} \\
$z^{13}$: \action{$<$Set[Object]:Set[Object]$>$}{square} \\
\end{minipage}
}
}
\caption{\label{fig:example-prog}
Gold program actions for the utterance \utterance{There is one box with at least 2 yellow squares} according to our proposed logical language. The grammar-constrained decoder outputs a linearized abstract-syntax tree of the program in an in-order traversal.
}
\end{figure*}

\section{Dataset details}
\label{app:dataset}
To discover related utterance pairs within the NLVR dataset, we manually identify 11 sets of phrases that commonly occur in NLVR and can be interpreted in the same manner: 
\begin{enumerate}
\item \{ COLOR block at the base, the base is COLOR \}
\item \{ COLOR block at the top, the top is COLOR \}
\item \{ COLOR1 object above a COLOR2 object \}
\item \{ COLOR1 block on a COLOR2 block, COLOR1 block over a COLOR2 block \}
\item \{ a COLOR tower \}
\item \{ there is one tower, there is only one tower, there is one box, there is only one box \}
\item \{ there are exactly NUMBER towers, there are exactly NUMBER boxes \}
\item \{ NUMBER different colors \}
\item \{ with NUMBER COLOR items, with NUMBER COLOR blocks, with NUMBER COLOR objects \}
\item \{ at least NUMBER COLOR items, at least NUMBER COLOR blocks, at least NUMBER COLOR objects \}
\item \{ with NUMBER COLOR SHAPE, are NUMBER COLOR SHAPE, with only NUMBER COLOR SHAPE, are only NUMBER COLOR SHAPE \}
\end{enumerate}
In each phrase, we replace the abstract COLOR, NUMBER, SHAPE token with all possible options from the NLVR dataset to create grounded phrases. For example, \utterance{black block at the top}, \utterance{yellow object above a blue object}.
For each set of equivalent grounded phrases, we identify the set of utterances that contains any of the phrase. For each utterance in that set, we pair it with 1 randomly chosen utterance from that set. Overall, we identify related utterances for 1420 utterances (out of 3163) and make 1579 pairings in total; if an utterance contains two phrases of interest, it can be paired with more than 1 utterance.

\end{document}